# Implementation of Training Convolutional Neural Networks

Tianyi Liu, Shuangsang Fang, Yuehui Zhao, Peng Wang, Jun Zhang
University of Chinese Academy of Sciences, Beijing, China
{liutianyi14@mails.ucas.ac.cn}

## ABSTRACT

Deep learning refers to the shining branch of machine learning that is based on learning levels of representations. Convolutional Neural Networks (CNN) is one kind of deep neural network. It can study concurrently. In this article, we gave a detailed analysis of the process of CNN algorithm both the forward process and back propagation. Then we applied the particular convolutional neural network to implement the typical face recognition problem by java. Then, a parallel strategy was proposed in section4. In addition, by measuring the actual time of forward and backward computing, we analysed the maximal speed up and parallel efficiency theoretically.

## 1. INTRODUTION

Deep learning refers to a subfield of machine learning that is based on learning levels of representations, corresponding to a hierarchy of features, factors or concepts, where higher-lever concepts are defined from lower-lever ones, and the same lower-lever concepts can help to define many higher-lever concepts.

Deep learning is learning multiple levels of representation and abstraction, helps to understand the data such as images, audio and text. The concept of Deep Learning comes from the study of Artificial Neural Network, Multilayer Perceptron which contains more hidden layers is a Deep Learning structure.

In the late 1980s, the invention of Back Propagation algorithm used in Artificial Neural Network brings hope to machine learning and creates a trend of machine learning based on statistical models. In the 1990s, a variety of Shallow Learning models have been proposed such as Support Vector Machines (SVM), Boosting, Logistic Regression (LR). The structure of these models can be seen as one hidden node (SVM, Boosting), or no hidden nodes (LR). These models gained a great success both in theoretical analysis and applications.

In 2006, Geoffrey Hinton who is the professor of University of Toronto, Canada and the dean of machine learning and his students Ruslan Salakhutdinov published an article in "Science", led to a trend of machine learning in academia and industry. The article had two points: 1) Artificial Neural Network with multiple hidden layers has an excellent ability of characteristic learning. The characteristics obtained from learning have an essential description to data, then facilitate visualization or classification. 2) The difficulties of deep neural network in training can overcome by layer-wise pre-training. In this article, the implementation of layer-wise pre-training is achieved through unsupervised learning.

Feedforward neural network or Multilayer Perceptron with multiple hidden layers in artificial neural networks is usually known as Deep Neural Networks (DNNs). Convolutional Neural Networks (CNN) is one kind of feedforward neural network. In 1960s, when Hubel and Wiesel researched the neurons used for local sensitive orientation-selective in the cat's visual system, they found the special network structure can effectively reduce the complexity of Feedback Neural Networks and then proposed Convolution Neural Network. CNN is an efficient recognition algorithm which is widely used in pattern recognition and image processing. It has many features such as simple structure, less training parameters and adaptability. It has become a hot topic in voice analysis and image recognition. Its weight shared network structure make it more similar to biological neural networks. It reduces the complexity of the network model and the number of weights.

Generally, the structure of CNN includes two layers one is feature extraction layer, the input of each neuron is connected to the local receptive fields of the previous layer, and extracts the local feature. Once the local features is extracted, the positional relationship between it and other features also will be determined. The other is feature map layer, each computing layer of the network is composed of a plurality of feature map. Every feature map is a plane, the weight of the neurons in the plane are equal. The structure of feature map uses the sigmoid function as activation function of the convolution network, which makes the feature map have shift invariance. Besides, since the neurons in the same mapping plane share weight, the number of free parameters of the network is reduced. Each convolution layer in the convolution neural network is followed by a computing layer which is used to calculate the local average and the second extract, this unique two feature extraction structure reduces the resolution.

CNN is mainly used to identify displacement, zoom and other forms of distorting invariance of two-dimensional graphics. Since the feature detection layer of CNN learns by training data, it avoids explicit feature extraction and implicitly learns from the training data when we use CNN. Furthermore, the neurons in the same feature map plane have the identical weight, so the network can study concurrently. This is a major advantage of the convolution network with respect to the neuronal network connected to each other. Because of the special structure of the CNN's local shared weights makes it have a unique advantage in speech recognition and image processing. Its layout is closer to the actual biological neural network. Shared weights reduces the complexity of the network. In particular multi-dimensional input vector image can directly enter the network, which avoids the complexity of data reconstruction in feature extraction and classification process.

Face recognition is a biometric identification technology based on the facial features of persons. The study of face recognition system began in the 1960s, in the late 1980s with the development of computer technology and optical imaging techniques it has been improved; in the late 1990s it truly entered the stages of initial applications. In practical applications, such as monitoring system, the collected face images captured by cameras are often low resolution and with great pose variations. Affected by pose variation and low resolution, the performance of face recognition degrades sharply. And pose variations bring great challenge to face recognition. They bring nonlinear factors into face recognition. And some of the existing machine learning method mostly use shallow structure. Deep learning can achieve the approximation of complex function by a deep nonlinear network structure.

In this article, we use convolution neural network to solve face recognition. It can overcome the influence of pose or resolution in face recognition. Due to the long training time and the large recognition computing, it is difficult to meet the real-time requirements, or the delay exceeds the range of tolerance. So we use the cloud platform to concurrently speed up the computing process.

## 2. BACKGROUND AND RELATED WORK

Convolutional Neural Networks can be applied in many different areas. Yann LeCun and his team specially designed Convoutional Neural Networks to deal with the variability of 2D shapes, which are shown to outperform all other techniques.[1] Dan C and his team present a fast, fully parameterizable GPU implementation of Convolutional Neural Network variants for Image Classification. [2] Another team proposes two novel frontends for robust language identification (LID) using a convolutional neural network (CNN) trained for automatic speech recognition (ASR). [5] What's more, Convolutional Neural Networks are used in Visual Recognition[9] and many other areas, such as Facial Point Detection[6], House Numbers Digit Classification[10], Multi-digit Number Recognition from Street View Imagery[11].

Besides these works, many teams are focusing on the speed up of ConvNets. For example, Multi-GPU Training of ConvNets. In this work , Facebook AI Group consider a standard architecture [1] trained on the Imagenet dataset [2] for classification and investigate methods to speed convergence by parallelizing training across multiple GPUs.[4]

## 3. PRINCIPLE OF CONVELUTIONAL NEURAL NETWORKS

### 3.1 Methodology

Convolution neural network algorithm is a multilayer perceptron that is the special design for identification of two-dimensional image information . Always has more layers: input layer, convolution layer, sample layer and output layer. In addition, in a deep network architecturethe convolution layer and sample layer can have multiple. CNN is not as restricted boltzmann machine, need to be before and after the layer of neurons in the adjacent layer for all connections, convolution neural network algorithms, each neuron don't need to do feel global image, just feel the local area of the image. In addition, each neuron parameter is set to the same, namely, the sharing of weights , namely each neuron with the same convolution kernels to deconvolution image.

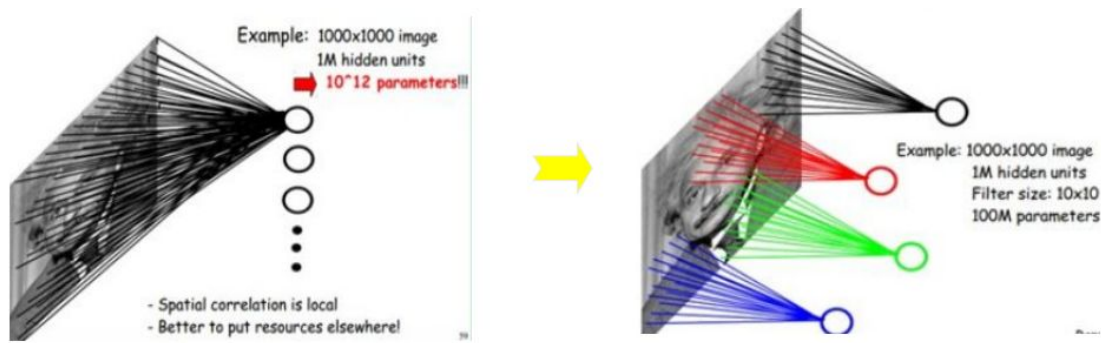

Fig 1 fully connection vs partial connection

CNN algorithm has two main processes: convolution and sampling .

Convolution process: use a trainable filter Fx, deconvolution of the input image (the first stage is the input image, the input  of the  after convolution  is the feature image of each layer, namely Feature Map), then add a bias bx, we can get convolution layer Cx.

A sampling process: n pixels of each neighborhood through pooling steps, become a pixel, and then by scalar weighting Wx + 1 weighted, add bias bx + 1, and then by an activation function, produce a narrow n times feature map Sx + 1.

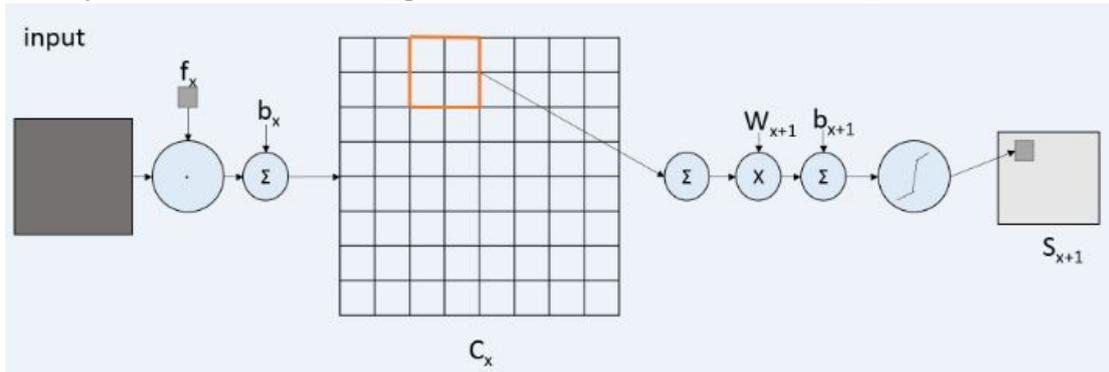

Fig 2 main process of CNN

The key technology of CNN is the local receptive field,  sharing of  weights , sub sampling by time or space, so as to extract feature and reduce the size of the training parameters.The advantage of CNN algorithm is that to avoid the explicit feature extraction, and implicitly to learn from the training data;The same neuron weights on the surface of the feature mapping, thus network can learn parallelly , reduce the complexity of the network;Adopting sub sampling structure by time or space, can achieve some degree of robustness, scale and deformation displacement;Input information and network topology can be a very good match, It has unique advantages in speech recognition and image processing.

## 3.2 CNN Architecture Design

CNN algorithm need experience in architecture design, and need to debug unceasingly in the practical application, in order to obtain the most suitable for a particular application architecture of CNN. Based on gray image as the input of 96 × 96, in the preprocess stage, turning it into 32 × 32 of the size of the image. Design depth of the layer 7 convolution model: input layer, convolution layer C1, sub sampling layer S1, convolution layer   C2, sampling layer S2, hidden layer H and output layer F.

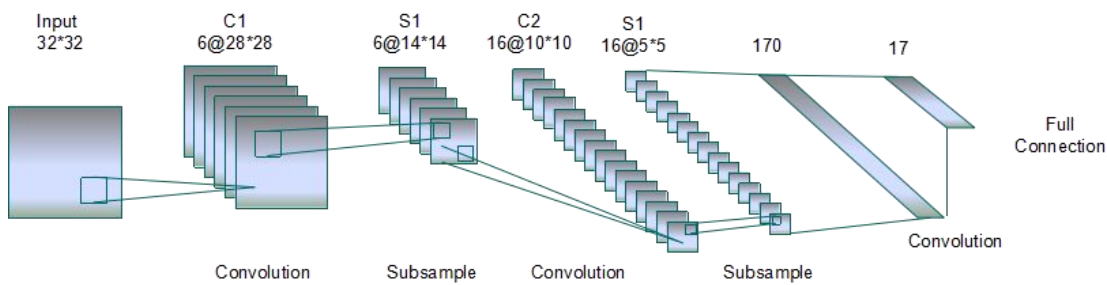

Fig 3 architecture of CNN in training faces

In view of the 32 × 32 input after preprocessing, There is a total of 17 different pictures.

C1 layer for convolution, convolution layer adopts 6 convolution kernels, each the size of the convolution kernels is 5 × 5, can produce six feature map, each feature map contains (32-5 + 1) × (32-5 + 1) = 28 × 28 = 784 neurons. At this point, a total of 6 × (5 × 5 + 1) = 156 parameters to be trained .

S1 layer for sub sampling, contains six feature map, each feature map contains 14 * 14 = 196 neurons. the sub sampling window is 2 × 2 matrix, sub sampling step size is 1, so the S1 layer contains 6 × 196 × (2 × 2 + 1) = 5880 connections. Every feature map in the S1 layer contains a weights and bias, so a total of 12 parameters can be trained in S1 layer .

C2 is convolution layer, containing 16 feature graph, each feature graph contains (14-5 + 1) (14-5 + 1) = 100 neurons and adopts full connection, namely each characteristic figure used to belong to own 6 convolution kernels with six characteristics of the sample layer S1 convolution and figure. Each feature graph contains 6 × 5 × 5 = 150 weights and a bias. So, C2 layer contains a total of 16 × (150 + 1) = 150 parameters to be trained.

S2 is sub sampling layer, containing 16 feature map, each feature map contains 5 × 5 neurons, S2 total containing 25 × 16 = 400 neurons. S2 on characteristic figure of sub sampling window for 2 × 2, so there is 32 trainable S2 parameters.

As a whole connection layer, hidden layer H contains 170 neurons, each neuron is connected to 400 neurons on S2. So H layer contains 170 × (400 + 1) = 48120 parameters feature map.

Output layer F for all connections, including 17 neurons. A total of 17 × (170 + 1) = 2907 parameters to be trained.

### 3.3 CNN Algorithm and Back propagation algorithm

3.1.1 Forward pass

output of neuron of row $k$, column $y$ in the $l$ th convolution layer and $k$ th feature pattern:

$$O_{x,y}^{(l,k)} = \tanh(\sum_{t=0}^{f-1}\sum_{r=0}^{k_h}\sum_{c=0}^{k_w} W_{(r,c)}^{(k,t)} O_{(x+r,x+c)}^{(l-1,t)} + Bias^{(l,k)}) \quad (3.1)$$

among them, $f$ is the number of convolution cores in a feature pattern。

output of neuron of row $x$, column $y$ in the $l$ th sub sample layer and $k$ th feature pattern:

$$O_{x,y}^{(l,k)} = \tanh(W^{(k)} \sum_{r=0}^{S_h}\sum_{c=0}^{S_w} O_{(x \times S_h + r, y \times S_w + c)}^{(l-1,k)} + Bias^{(l,k)}) \quad (3.2)$$

the output of the $j$ th neuron in $l$ th hide layer $H$:

$$O_{(l,j)} = \tanh(\sum_{k=0}^{s-1}\sum_{x=0}^{S_h}\sum_{y=0}^{S_w} W_{(x,y)}^{(j,k)} O_{(x,y)}^{(l-1,k)} + Bias^{(l,j)}) \quad (3.3)$$

among them, $s$ is the number of feature patterns in sample layer.

output of the $i$ th neuron $l$ th output layer $F$

$$O_{(l,i)} = \tanh(\sum_{j=0}^{H} O_{(l-1,j)} W^l_{(i,j)} + Bias^{(l,i)}) \qquad (3.4)$$

3.1.2 Back propagation

output deviation of the $k$ th neuron in output layer $O$:

$$d(O_k^O) = y_k - t_k \qquad (3.5)$$

input deviation of the $k$ th neuron in output layer:

$$d(I_k^O) = (y_k - t_k)\varphi'(v_k) = \varphi'(v_k)d(O_k^o) \qquad (3.6)$$

weight and bias variation of $k$ th neuron in output $O$:

$$\Delta W_{k,x}^O = d(I_k^O) y_{k,x} \qquad (3.7)$$

$$\Delta Bias_k^O = d(I_k^O) \qquad (3.8)$$

output bias of $k$ th neuron in hide layer $H$:

$$d(O_k^H) = \sum_{i=0}^{i<17} d(I_i^O) W_{i,k} \qquad (3.9)$$

input bias of $k$ th neuron in hide layer $H$:

$$d(I_k^H) = \varphi'(v_k)d(O_k^H) \qquad (3.10)$$

weight and bias variation in row $x$, column $y$ in the $m$ th feature pattern, a former layer in front of $k$ neurons in hide layer $H$

$$\Delta W_{m,x,y}^{H,k} = d(I_k^H) y_{x,y}^m \qquad (3.11)$$

$$\Delta Bias_k^H = d(I_k^H) \qquad (3.12)$$

output bias of row $x$, column $y$ in $m$ th feature pattern, sub sample layer $S$

$$d(O_{x,y}^{S,m}) = \sum_{k}^{170} d(I_{m,x,y}^H) W_{m,x,y}^{H,k} \qquad (3.13)$$

input bias of row $x$, column $y$ in $m$ th feature pattern, sub sample layer S

$$d(I_{x,y}^{S,m}) = \varphi'(v_k)d(O_{x,y}^{S,m}) \qquad (3.14)$$

weight and bias variation of row $x$, column $y$ in $m$ th feature pattern, sub sample layer $S$

$$\Delta W^{S,m} = \sum_{x=0}^{fh} \sum_{y=0}^{fw} d(I_{\lfloor x/2 \rfloor, \lfloor y/2 \rfloor}^{S,m}) O_{x,y}^{C,m} \qquad (3.15)$$

among them, $C$ represents convolution layer.

$$\Delta Bias^{S,m} = \sum_{x=0}^{fh}\sum_{y=0}^{fw} d(O_{x,y}^{S,m}) \quad (3.16)$$

output bias of row $x$, column $y$ in $k$ th feature patter, convolution layer $C$

$$d(O_{x,y}^{C,k}) = d(I_{\lfloor x/2 \rfloor, \lfloor y/2 \rfloor}^{S,k})W^k \quad (3.17)$$

iutput bias of row $x$, column $y$ in $k$ th feature patter, convolution layer $C$

$$d(I_{x,y}^{C,k}) = \varphi'(v_k)d(O_{x,y}^{C,k}) \quad (3.18)$$

weight variation of row $r$, column $c$ in $m$ th convolution core, corresponding to $k$ th feature pattern in $l$ th layer, convolution $C$.

$$\Delta W_{r,c}^{k,m} = \sum_{x=0}^{fh}\sum_{y=0}^{fw} d(I_{x,y}^{C,k})O_{x+r,y+c}^{l-1,m} \quad (3.19)$$

total bias variation of the convolution core

$$\Delta Bias^{C,k} = \sum_{x=0}^{fh}\sum_{y=0}^{fw} d(I_{x,y}^{C,k}) \quad (3.20)$$

## 4. EXPERIMENTS

### 4.1 Setups

Table I setups of the experiment

| CPU | Intel Core i5-3570 CPU3.40GHz×4 |
|---|---|
| The number of CPU cores | 4 |
| Memory Size | 4G |
| Operation System | Ubuntu 14.10 |

### 4.2 Parallel Strategy and Parallel Efficiency

This analysis is based on a hypothesis that both serial and parallel method have the same number of training  In serial realization method, the total execution time is N times of the sum of t1 and t2

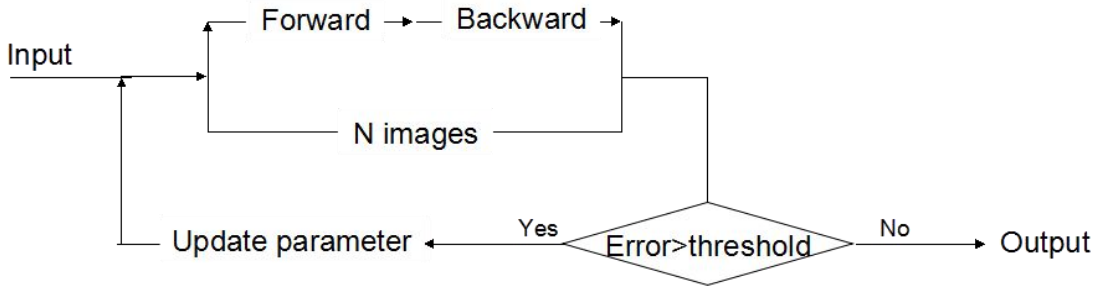

Fig 4 the serial algorithm of training

Time of serial execution: $(t_1 + t_2) \times N + t_3 = t_{serial}$

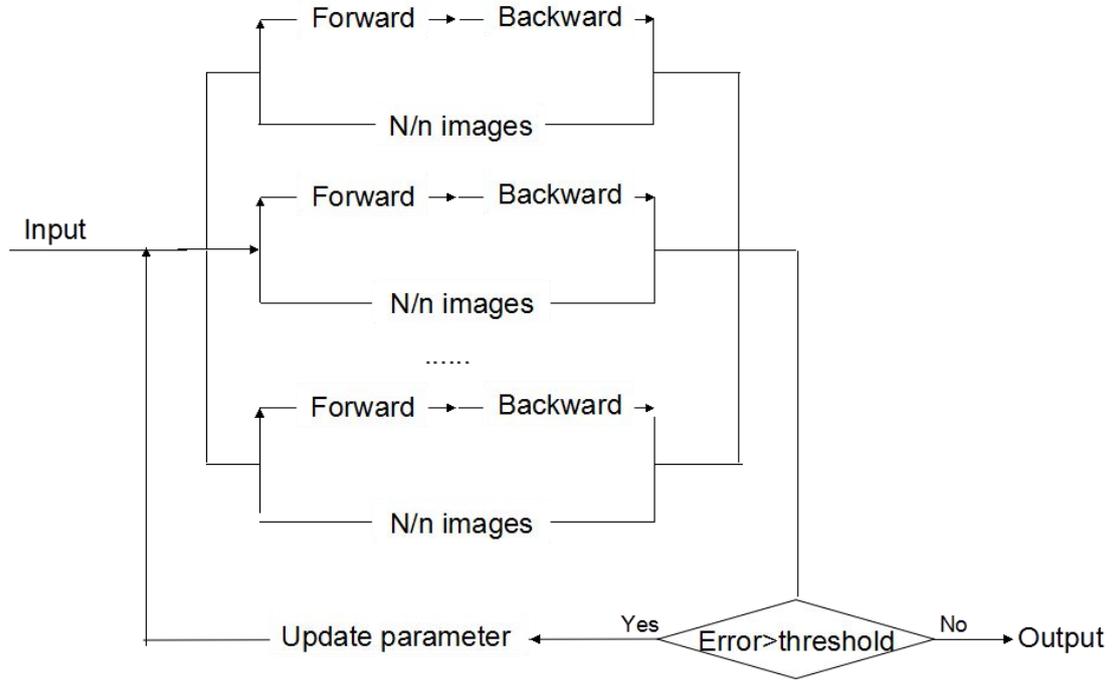

Fig 5 the parallel strategy

Time of parallel execution: $\max\{(t_1' + t_2')\} \times (N/n) + t_3' = t_{parallel}$

speed-up ratio $= t_{serical} / t_{parallel}$

Speed-up efficiency=speed-up ratio/n
N: num of images
n: num of nodes
$t_1$ :time of forward pass for training a picture
$t_2$ :time of backward propagation for training a picture
$t_3$ :time for updating weight and bias of convolution neural network

### 4.3 Results Analysis

The data set we used is from Yale Face Database. We choose 136 images to analysis. When we run our algorithm we need to divide it into two phases. First, we need to train our algorithm. The purpose of this phase is to determine the minimum error which will be used in the next phase. So we must ensure that the algorithm can converge at a certain point. During the training process, the error will be reduced until it becomes a constant. The constant will be used as a threshold in the next step. Figure 1 shows the error will not change after repeating 4 times. So the best error is 4. The horizontal axis represents the number of iterations, and the vertical axis represents error.

Seconds, we can use the constant obtained from the first step as the threshold to judge whether the algorithm can stop. Table II shows when the training process is successful the time consumed by the algorithm. In the table, "yes" represents the algorithm is succeed, in contrast "no" represents the algorithm is failed. We can see that the average time used by the algorithm is

12374.3 milliseconds. The reason why the algorithm is failed is it fell into the local optimum.

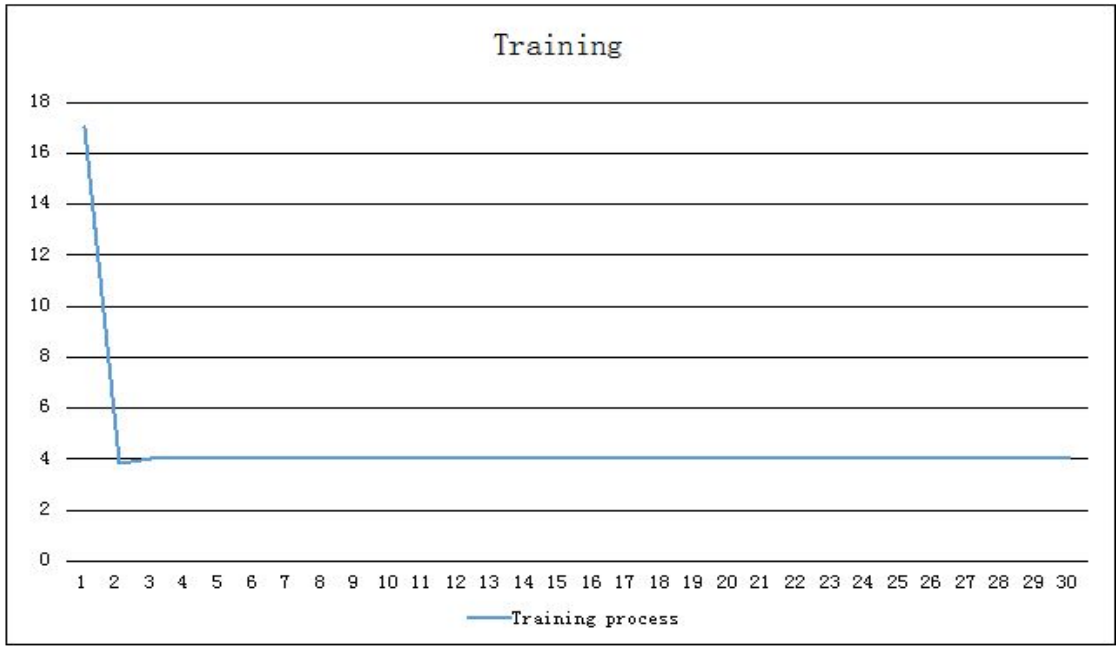

Fig 6 the process of training

Table II time consumed by the algorithm

|  | 1 | 2 | 3 | 4 | 5 | 6 | 7 | 8 | 9 | 10 | 11 |
|---|---|---|---|---|---|---|---|---|---|---|---|
| result | yes | yes | yes | no | yes | no | yes | yes | yes | yes | yes |
| time/ms | 11642 | 23163 | 11615 |  | 11794 | 11715 | 11966 | 11687 | 11633 | 11892 | 11636 |
| average/ms | 12874.3 | | | | | | | | | | |

### 4.4 Theoretically Analysis of The Maximal Speedup

By measuring the time overhead of training, especially the average time of $t_1$, $t_2$ and $t_3$, the maximal speedup and speed-up efficiency are listed below:

Time of serial execution: 5317.000000
Time of parallel execution :2665.000000
speed-up ratio         : 1.995122
Speed-up efficiency    : 0.997561

## 5. CONCLUSION

In this work, we accomplish face recognition by using deep learning algorithm. We mainly apply the algorithm of convolution neural network to excavate the deep information of multi-layer network in the process of face recognition .And we also utilize the algorithm to make parallel computing on the cloud platform for accelerating the process of face recognition, analyzing theoretical acceleration ratio, and experimental verification. Experimental results show that we have achieved good results. Of course, the parallelism we do is coarse-grained, and there are still many modules that can be fine-grained in the algorithm. This will be the focus for us in the future to continue to improve the work.

## ACKNOWLEDGMENT

During the time we work together to complete the course task, Prof. Chen takes much effort to offer us guide and help. So the first person we must offer our thanks to is Mr.Chen. At the same time, our major team leader Tianyi Liu, also deserves our sincerest thanks. He has done much work to organize teammates and coordinate everyone's work. And lastly,thanks to everybody in our team, we reach a consensus and we make concerted efforts, then we can complete our work in time and publish it on the website .